AGMN: Association Graph-based Graph Matching Network for Coronary Artery Semantic Labeling on Invasive Coronary Angiograms

Chen Zhao[1], Zhihui Xu[2], Jingfeng Jiang[3], Michele Esposito[4], Drew Pienta[5], Guang-Uei Hung[6], Weihua Zhou[1,7*]

1. Department of Applied Computing, Michigan Technological University, Houghton, MI, USA

2. Department of Cardiology, The First Affiliated Hospital of Nanjing Medical University, Nanjing, China

3. Department of Biomedical Engineering, Michigan Technological University, Houghton, MI, USA

4. Department of Cardiology, Medical University of South Carolina, Charleston, SC, USA

5. Mechanical Engineering-Engineering Mechanics, Michigan Technological University, Houghton, MI, USA

6. Department of Nuclear Medicine, Chang Bing Show Chwan Memorial Hospital, Changhua, Taiwan

7. Center for Biocomputing and Digital Health, Institute of Computing and Cyber-systems, and Health Research Institute, Michigan Technological University, Houghton, MI, USA

Corresponding author:

Weihua Zhou, PhD, Tel: +1 906-487-2666

E-mail address: whzhou@mtu.edu

Mailing address: 1400 Townsend Dr, Houghton, MI 49931

**Abstract**

Semantic labeling of coronary arterial segments in invasive coronary angiography (ICA) is important for automated assessment and report generation of coronary artery stenosis in the computer-aided diagnosis of coronary artery disease (CAD). However, separating and identifying individual coronary arterial segments is challenging because morphological similarities of different branches on the coronary arterial tree and human-to-human variabilities exist. Inspired by the training procedure of interventional cardiologists for interpreting the structure of coronary arteries, we propose an association graph-based graph matching network (AGMN) for coronary arterial semantic labeling. We first extract the vascular tree from invasive coronary angiography (ICA) and convert it into multiple individual graphs. Then, an association graph is constructed from two individual graphs where each vertex represents the relationship between two arterial segments. Thus, we convert the arterial segment labeling task into a vertex classification task; ultimately, the semantic artery labeling becomes equivalent to identifying the artery-to-artery correspondence on graphs. More specifically, using the association graph, the AGMN extracts the vertex features by the embedding module, aggregates the features from adjacent vertices and edges by graph convolution network, and decodes the features to generate the semantic mappings between arteries. By learning the mapping of arterial branches between two individual graphs, the unlabeled arterial segments are classified by the labeled segments to achieve semantic labeling. A dataset containing 263 ICAs was employed to train and validate the proposed model, and a five-fold cross-validation scheme was performed. Our AGMN model achieved an average accuracy of 0.8264, an average precision of 0.8276, an average recall of 0.8264, and an average F1-score of 0.8262, which significantly outperformed existing coronary artery semantic labeling methods. In conclusion, we have developed and validated a new algorithm with high accuracy, interpretability, and robustness for coronary artery semantic labeling on ICAs.

**Keywords:** coronary artery disease, invasive coronary angiography, coronary arterial anatomy, semantic labeling, graph matching network

## 1. Introduction

Coronary artery disease (CAD), caused by narrowing or blockages of the coronary arteries, is the most common cardiovascular disease in the United States [1,2]. The narrowing is due to the buildup of fatty plaque along the artery walls, composed of cholesterol, lipids, and fibrous tissue [3]. If one or more of these arteries become severely obstructed, thereby reducing downstream blood flow, this may have the deleterious consequence of resulting in myocardial ischemia or infarction [4].

Invasive coronary angiography (ICA) remains the gold standard for diagnosing CAD [5]. ICA involves the injection of contrast media into the epicardial arteries with the acquisition of continuous fluoroscopy angiograms. For clinical decision-making, a trained cardiologist diagnoses CAD by subjectively assessing the percent stenosis, or narrowing, of diseased arterial segments compared to normal arterial segments. Automatic labeling of anatomical branches provides critical information for the diagnosis, report generation, and region of interest visualization [6]. Meanwhile, measurement of coronary artery stenosis influences patient management while improving diagnostic efficiency and confidence [7]. In clinical practice, cardiologists and radiologists report the pathological findings for each arterial segment according to the American Heart Association guidelines [8].

The coronary vascular tree is complex and contains two major systems: the left coronary artery (LCA) and the right coronary artery (RCA) systems. The LCA is more clinically relevant, given it provides most of the blood supply to the left ventricle and is therefore associated with higher in-hospital mortality among patients undergoing percutaneous coronary intervention [9]. The LCA system contains the left main artery (LMA), left descending artery (LAD), and left circumflex artery (LCX), as well as multiple diagonal arteries (D) and obtuse marginal (OM) branches [10]. The LMA arises from the aorta above the left cusp of the aortic valve and perfuses the left ventricle's anterior, septal, and lateral walls. As shown in Figure 1(a), the LMA bifurcates into two main branches: the left anterior descending artery (LAD), which courses between the left and right ventricles towards the apex along the anterior interventricular sulcus, and the left circumflex artery (LCX), which runs laterally along the atrioventricular groove. The D and OM branches originate from the LAD and LCX, respectively. It is worth noting that the main branches of the coronary arterial tree are the LMA, LAD, and LCX, while the D and OM arteries are the side branches. Compared to brain vessels and airways, semantic labeling of coronary arteries is more challenging due to variations in length, size, and position [10,11].

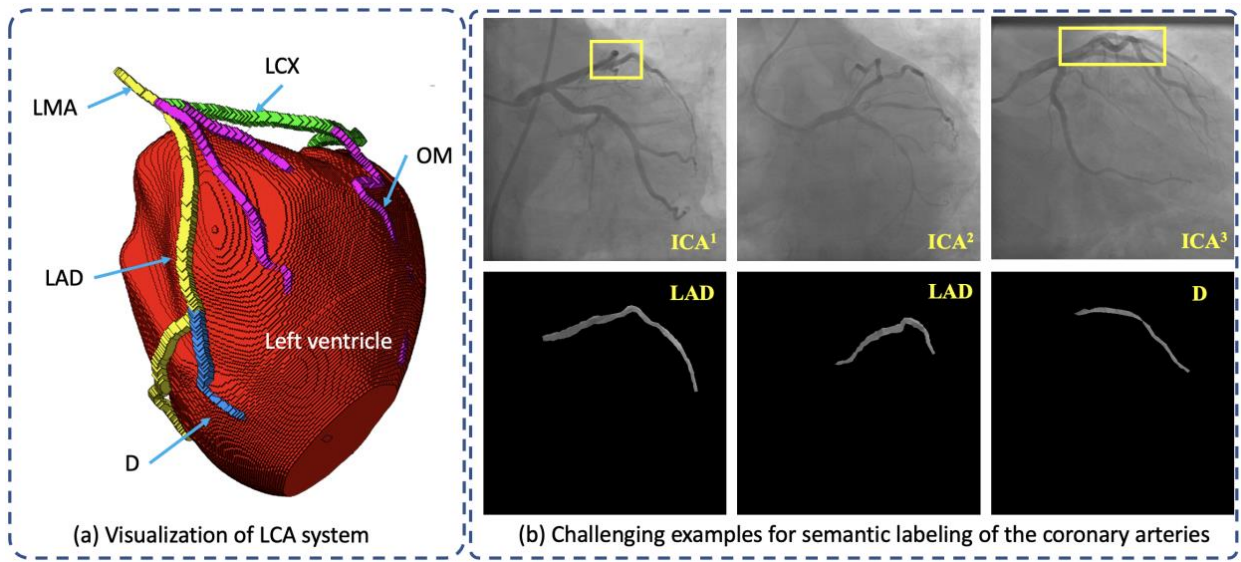

**Figure 1**. (a) Visualization of coronary arterial tree in LCA system along with left ventricle and (b) challenging examples for coronary artery semantic labeling. The yellow rectangles indicate the overlay of arteries in 2D. For these three examples in (b), the LAD branches from the first two ICAs showed similarity

in morphological and pixel-wise features with the D branch of the third ICA, demonstrating some challenges accociated with coronary semantic segmentation.

Pixel-intensity-based models have difficulties distinguishing each arterial segment and generating semantic segmentation because of the morphological similarity among different branches in the coronary vascular tree and the overlap of the arteries in 2D (projection) ICA, as shown in Figure 1 (b). Also, the coronary arteries not only span over a long distance but also show similar semantic features with each other [12], making it challenging to associate them with the exact branches. False identifications of arterial segments not only impair the understanding of the structure of the arterial tree but also causes inaccurate assessments of vascular stenosis in the clinical workflow [13]. Existing methods only relying on position and imaging features may produce unsatisfactory results when processing complicated coronary vasculature [14]. The topology is a crucial factor in arterial identification, inspiring us to convert arteries and their connectivity into graphs and perform coronary artery semantic labeling using graphs.

In this paper, we propose a novel algorithm to perform coronary artery semantic labeling using ICAs. The key innovation of our coronary artery semantic labeling method is the use of a graph-matching network (GMN) to build the semantic correspondence between arterial segments from different ICAs. Formally, coronary artery semantic labeling is equivalent to finding the corresponding one-to-one mapping for arterial segments from two different arterial graphs generated from the coronary vascular trees. This study proposes an association-graph-based graph matching network (AGMN) to learn the similarities between arterial segments from two (arterial) individual graphs. Consequently, the labeled segments classify the unlabeled arterial segments to achieve semantic labeling by learning the matching of arterial branches between two individual graphs.

The workflow of the proposed graph-matching approach for coronary artery semantic labeling is shown in Figure 2. We first performed the coronary artery tree extraction as preprocessing the original ICA image (see Figure 2 (a)). The entire vascular tree is extracted using FP-U-Net++, a coronary artery binary segmentation model from a prior publication [15]. Then each centerline is extracted, and key points within the centerline are detected. Each set of centerlines, including its affiliated key points, is further simplified and denoised by applying a set of designated rules. Collectively, the vascular tree can be converted into an individual graph.

For each edge and node in the individual graph, we extract the pixel-wise, positional, and topological features for feature representation. An association graph is constructed from two individual graphs to convert the arterial segment labeling task into a vertex classification task where each vertex represents the relationship between two arterial segments, as shown in Figure 2 (b). The AGMN is then proposed to perform the vertex classification as well as graph matching and coronary artery semantic labeling.

Extensive experiments, including the comparative experiments established by deep-learning and traditional machine-learning methods for coronary artery semantic labeling, confirmed that the proposed AGMN quantitatively and qualitatively outperformed other state-of-the art methods in accuracy and robustness. The contributions and innovations of this work are as follows:

1) We designe five specific rules and a pipeline to preprocess the vascular tree and generate the individual graphs according to ICAs.
2) We convert the pixel-to-pixel semantic segmentation task into a graph-matching task for coronary artery semantic labeling. Our designed association graph-based graph-matching neural network significantly outperforms the existing coronary arterial semantic labeling methods.
3) We exploit the robustness of the proposed method for coronary artery semantic labeling using the corrupted datasets.

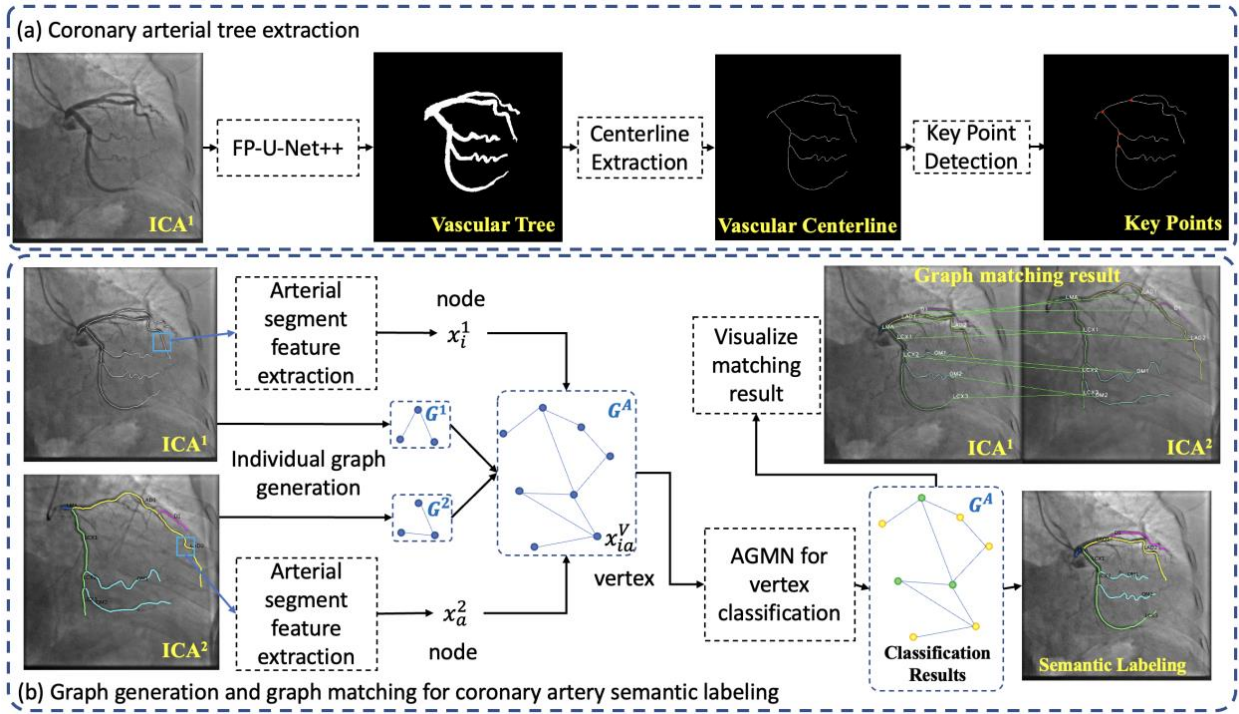

**Figure 2**. Workflow of the proposed coronary artery preprocessing and Association-graph-based Graph Matching Network (AGMN) for coronary artery semantic labeling. **(a)** Coronary arterial tree extraction and key point detection. **(b)** Graph generation and graph matching for coronary artery semantic labeling. Each ICA forms an individual graph consisting of nodal features extracted from each arterial segment. An association graph is built by considering the connectivity of those two individual graphs. The AGMN classifies vertices in the association graph into positive (in green) and negative vertices (in yellow). The positive vertex represents the corresponding nodes in the individual graphs that are matched and have identical semantic labels. Note that a node (i.e., an arterial segment) in the individual graph is referred to as a *node*, while a node in the association graph is referred to as a *vertex*.

## 2. Related Work

**Coronary artery semantic segmentation**. Coronary semantic segmentation is a challenging problem because coronary arteries show similar features in pixel-wise appearance and morphology. According to the problem definition, existing methods in the literature can be divided into two categories: 1) pixel-to-pixel-based image semantic segmentation approaches and 2) multi-class classification-based arterial segment labeling approaches.

The pixel-to-pixel-based image semantic segmentation model aims to assign a label to each pixel, a.k.a. pixel-level classification [16]. Xian et al. proposed an end-to-end attention residual U-Net for major artery segmentation on ICAs [17], where those major arteries include LAD, LXA, and RCA. Their model achieved an average F1-score > 0.8 among 89% of selected ICAs. Silva et al. designed an EfficientU-Net++ and achieved a Dice score of 0.8904 for major artery segmentation [18]. Zhang et al. proposed a progressive perception learning framework containing the context, interference, and boundary perception modules for major coronary artery segmentation. Although Zhang et al. achieved an average Dice score of greater than 0.95 for all types of major arteries [12], their model omitted the side branches, such as D and OM branches, limiting its clinical use.

The multi-class classification-based arterial labeling model is to assign a label to each arterial segment. Compared to the end-to-end pixel-level classification, this approach focuses on segment-level classification. Our previous work [14] is a machine-learning-based coronary artery semantic labeling method on ICAs, which extracted the hand-crafted features for each arterial segment and employed the support vector machine (SVM) for segment classification. Without using deep learning, the performance of feature embedding was limited and thus induced a low segment classification performance. Cao et al. employed prior knowledge and designed an iterative algorithm to label coronary arteries from main branches to side branches on 3D Cardiac CT angiograms (CCTA) [19]. Wu et al. proposed a bi-directional tree-based long short-term memory (LSTM) network for arterial segment semantic labeling on CCTAs [6]. The arterial spatial locations and directions were used as the features for arterial segment classification. Yang et al. converted the coronary artery semantic labeling task into a graph edge classification task and designed a partial-residual graph convolutional network (GCN) for artery classification on CCTAs [20]. However, the three methods mentioned above were developed for CCTA images instead of using ICA images. In short, the arterial segment classification method using ICA images is underdeveloped.

**Graph-matching using deep learning**. Graph-matching aiming to establish a meaningful node and edge correspondence between two graphs is often formulated as a graph edit distance problem [21], a maximum common subgraph problem [22], or a quadratic assignment problem [23]. However, all these solutions are NP-hard. Instead of solving the NP-hard problem, a graph-matching neural network is proposed to find the node correspondence between graphs. GCN has been proposed to process the non-Euclidean data to aggregate features from the adjacent nodes and edges [24]. Nowak et al. presented a GCN-based method for solving the quadratic assignment problem by learning the pre-defined affinity matrix [25]. Wang et al. proposed a cross-graph affinity learning approach for permutation learning to solve the graph-matching problem [26]. Specifically, in Wang et al.'s approach, the graph-matching problem was relaxed to the linear assignment problem, and a Sinkhorn classifier was adopted as the combinatorial solver. It is worth noting that all three approaches above used an exact matching criterion, requiring the number of nodes between graphs to be identical. However, the anatomical structures between coronary vascular trees from different subjects vary; thus, an inexact graph-matching algorithm is required for coronary artery semantic labeling [27].

## 3. Methodology

The approach presented in this study focuses on segment-level classification for coronary artery semantic labeling. Using AGMN, the problem of the semantic labeling task is converted into an equivalent problem of finding one-to-one or one-to-zero mapping for arterial segments from two different vascular trees. Figure 1 illustrates the overall workflow, followed by details in subsequent sections below.

### 3.1. Coronary arterial tree extraction and individual graph generation

Coronary arterial contours are extracted using our previously developed Feature Pyramid U-Net++ (FP-U-Net++) [15]. Then, each centerline is extracted to reduce redundant foreground pixels in a binary image while preserving the connectivity and topology of the vascular tree. As a result, the vascular tree's centerline and the arterial segments' diameters are calculated, as shown in Figure 2 (a). The workflow of individual graph generation is shown in Figure 3.

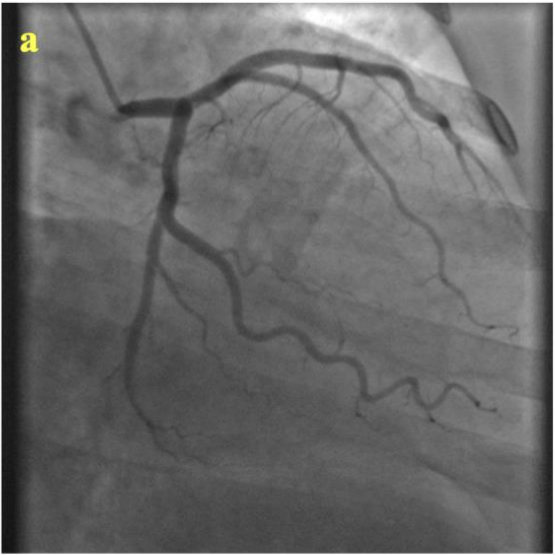
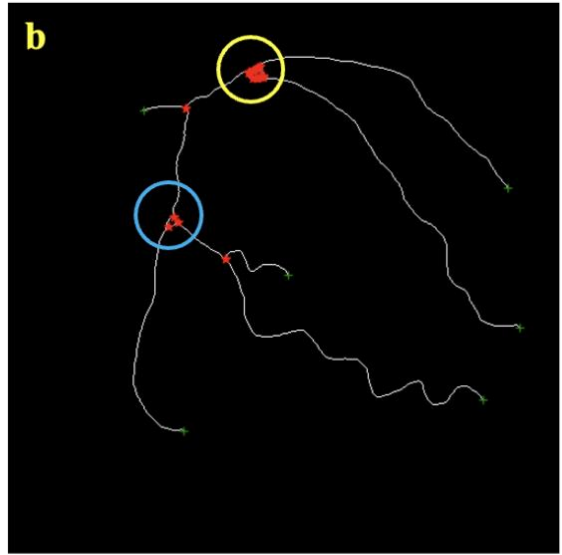
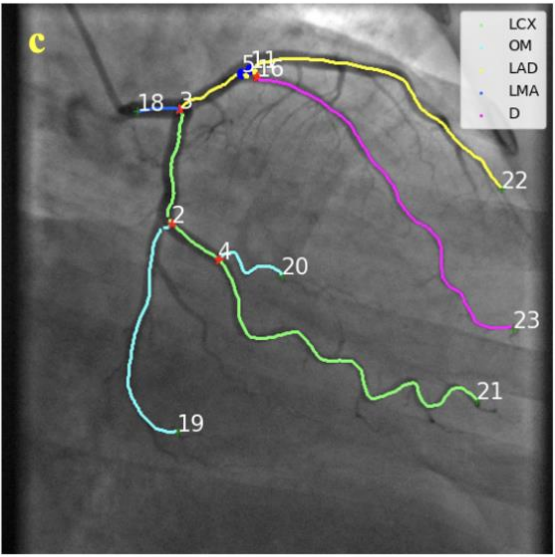
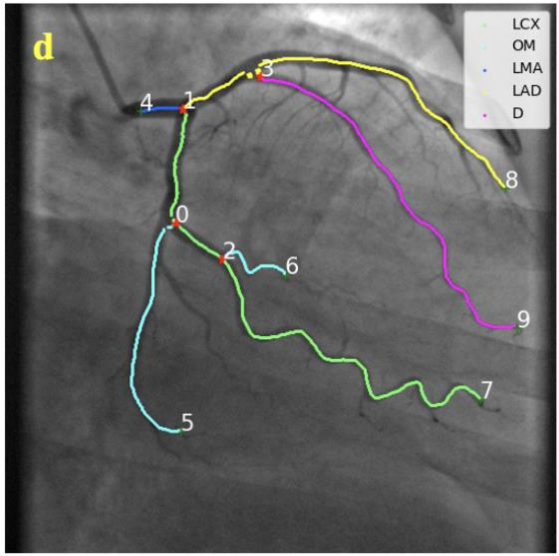
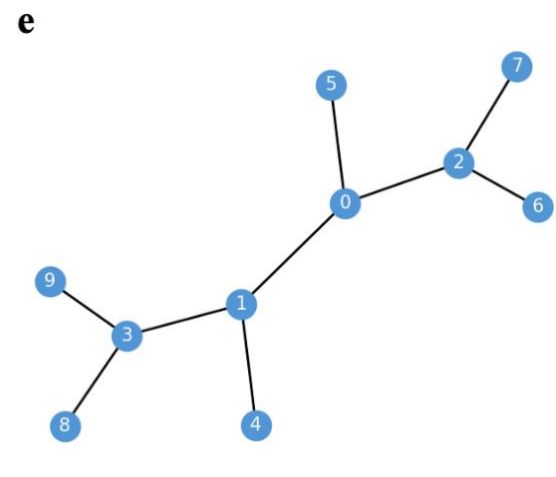
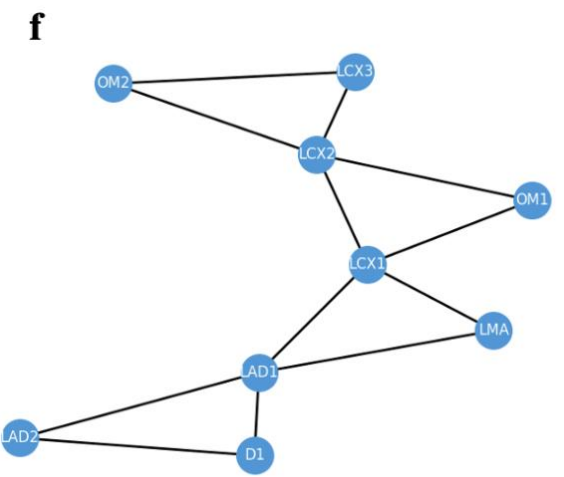

**Figure 3.** Workflow of individual graph generation and vascular centerline post-processing: **(a)** original ICA image; **(b)** centerline extraction and key point detection. The bifurcation points with degree > 3 are marked in red, and the endpoints with degree = 1 are marked in green; **(c)** vascular structure after merging the splitting points in the yellow region in (b) and deleting the cycle in the blue region in (b). In this sub-figure, the bifurcation points with degree > 3 are marked in red, the endpoints with degree = 1 are marked in green, and the degree two points generated by merging the splitting points are marked in blue; **(d)** vascular structure after merging the degree two points in (c); **(e)** generated individual graph; **(f)** the final individual graph by switching nodes and edges of the individual graph in (e). Since the graph matching is to find the node (arterial segment) correspondence rather than the edge correspondence, we switch node and edge in the individual graph in (e). In (c), (d), and (f), the classes of the arterial segments are labeled; in (c), (d), and (e), the node indices are annotated.

To convert the vascular tree into an arterial graph, we define two types of points: bifurcation points and endpoints. A bifurcation point is an intersection point that bifurcates the arterial segments into sub-segments. An endpoint represents the end of an arterial segment. This conversion process iterates through all points in a centerline until all bifurcations and endpoints are extracted. As shown in Figure 3 (b), each centerline contains multiple arterial segments and multiple bifurcations and endpoints. To find the links between one bifurcation point and one endpoint, or links between two bifurcation points, each bifurcation and endpoint is removed from the vascular centerline. Consequently, each vascular centerline is separated into several arterial segments.

We design several rules and an algorithm to eliminate errors and build the individual graph.

*i) Delete the capillary segments.* If the maximum diameter associated with a point in the centerline is smaller than the threshold $T_d$, then it has no significance for clinical analysis and is removed [15]. In addition, any short arterial segments with less than $T_c$ pixels in the centerline are also removed.

*ii) Merge splitting points.* Typical errors in an automatically generated arterial graph are induced by splitting points [28]. If two bifurcation points are located closely, then the splitting points affect graph topology creation, as shown in the yellow circle in Figure 3 (b). A threshold was set to remove two splitting points if the Euclidian distance between two bifurcation points is smaller than the threshold $T_{sp}$. As a result, those two points and related edges are merged.

*iii) Delete cycles.* To build an undirected acyclic graph for graph convolution, cycles in the graph must be deleted. If a cycle exists, the arterial segment with a smaller diameter is removed, as shown in Figure 3 (b).

*iv) Merge degree two points.* After removing the capillary arterial segments and the splitting points, the degrees of the bifurcation points are reduced to two. Then those bifurcation points are merged into the connected two arterial segments, as shown in Figure 3 (c) and (d).

*v) Switch nodes and edges.* The bifurcation points and endpoints are nodes in a graph, and the edges are the link between nodes. Naturally, each node in the individual graph represents a bifurcation point or an endpoint, and each edge represents an arterial segment. However, our graph-matching network aims at building node correspondence rather than edge correspondence. Semantic labeling requires the network to build relationships between arterial segments rather than the detected key points. Thus, we switch nodes and edges in the individual graph so that each node represents an arterial segment, and each edge represents a bifurcation node (with degree $\geq 3$). Consequently, a graph is generated for each vascular tree, denoted as $G = (V, E)$ where $V$ is the node set, and $E$ is the edge set. The generated graph is shown in Figure 3 (f). The adjacent matrix of the graph is generated by the connectivity of the vascular tree.

Our algorithm for arterial tree modification and individual graph generation is shown in Algorithm 1.

**Algorithm 1.** Individual graph generation from a vascular tree.

> **Input:**
> - $X$: input ICA image.
> - $T_d$: threshold for removing an arterial segment if its diameter is smaller than $T_d$.
> - $T_c$: threshold for removing an arterial segment if its centerline length is shorter than $T_c$.
> - $T_{sp}$: distance threshold for removing splitting points.
>
> **Output:** $G = (V, E)$: Generated individual graph.
> 1. Extract a vascular tree by FP-U-Net++ using $X$ and generate the vascular centerline;
> 2. Find bifurcation points and end points to separate the vascular centerline to arterial segments;
> 3. Calculate diameters and remove capillary segments by $T_d$;
> 4. Merge splitting points by $T_{sp}$;
> 5. Delete cycles if two arterial segments stem from the same point and end at the same point;
> 6. Merge degree two points if the degrees of splitting points are reduced into two;
> 7. Generate the individual graph and switch nodes and edges to create the final individual graph $G$.

### 3.2. Graph matching neural network for arterial segment labeling

Graph matching aims to find the node correspondence between two individual graphs $G^1 = (V^1, E^1)$ and $G^2 = (V^2, E^2)$, where $|V^1| = n_1$, $|V^2| = n_2$, $|E^1| = n_{e1}$ and $|E^2| = n_{e2}$. Without loss of generality, we assume $n_1 \leq n_2$ in this study. The two-graph matching problem can be written as a quadratic assignment programming (QAP) [29,30], defined as

$$J(X) = vec(M)^T K vec(M), s.t. M \in \{0,1\}^{n_1 \times n_2} \tag{1}$$

where $M \in \mathbb{R}^{n_1 \times n_2}$ is the permutation matrix encoding node-to-node correspondence and $K \in \mathbb{R}^{n_1 n_2 \times n_1 n_2}$ is the affinity matrix of the association graph $G^A = (V^A, E^A)$ generated by the connectivity of the graphs $G^1$ and $G^2$, where $|V^A| = n_1 \times n_2$ and $|E^A| = 2 \times n_{e1} \times n_{e2}$. $vec$ indicates the vectorization. To eliminate ambiguity, the node (an arterial segment) in the individual graph is defined as *node*, while the node in the association graph is defined as *vertex*.

The vertices of the association graph $V^A = V^1 \times V^2$ encode the node-to-node correspondence, denoted as $V_{i,a}^A = (V_i^1, V_a^2)$. The diagonal elements in $K$ represent the node-to-node matching level. For vertex generation, each candidate's correspondence $(V_i^1, V_a^2) \in V^1 \times V^2$ will be considered a vertex $V_{ia}^A$. The edge in the association graph is denoted as $E_{ia,jb}^A$, where $i,j \in \{1, \ldots, n_1\}$ and $a,b \in \{1, \ldots, n_2\}$. For edge generation, we build an edge between a pair of vertices $V_{ia}^A$ and $V_{jb}^A$ if and only if there are two edges in their own graphs that $(V_i^1, V_j^1) \in E^1$ and $(V_a^2, V_b^2) \in E^2$. Thus, the graph-matching problem is converted to classify vertices in the association graph $G^A$ into either positive vertices or negative vertices. For example, if the node $V_i^1 \in G^1$ and $V_a^2 \in G^2$ have the same semantic labels, then the vertex $V_{ia}^A \in G^A$ is a positive vertex, and vice versa. The overall association graph-based graph-matching network contains the following five modules, as shown in Figure 4.

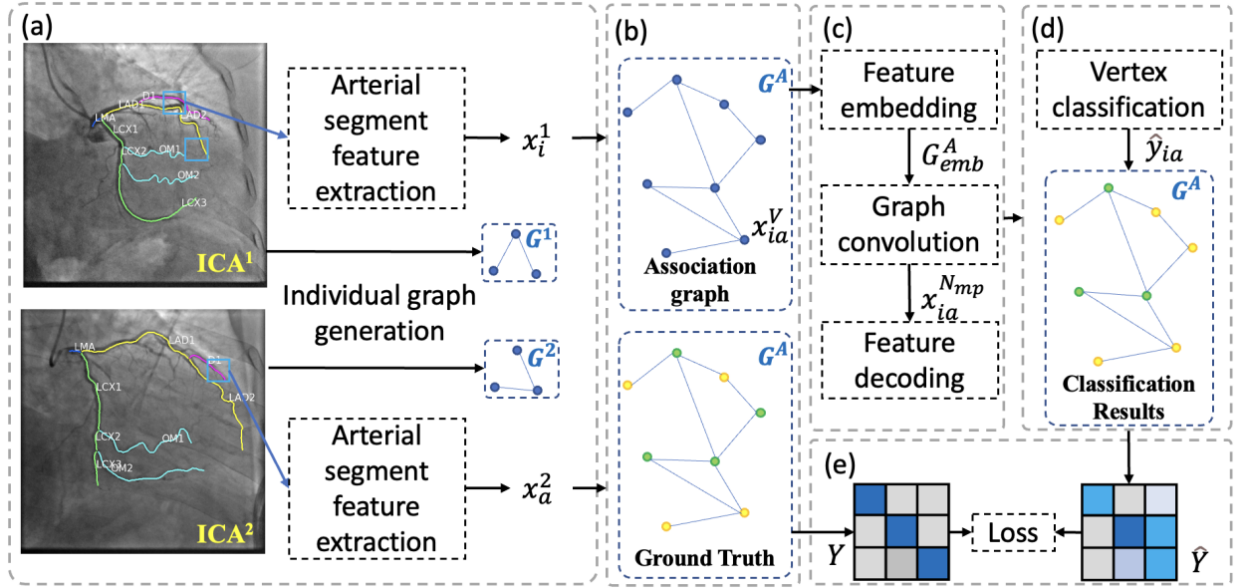

**Figure 4.** A flowchart showing the architecture of the association graph-based graph matching network: **(a)** individual graph generation and node feature extraction; **(b)** association graph generation, ground truth generation, and feature concatenation for vertices in the association graph; **(c)** feature representation learning, including feature embedding, graph convolution network and feature decoding; **(d)** vertex classification by major probability voting; **(e)** loss calculation and model training.

*i) Extracting features for nodes in individual graphs.* Using the processing algorithm, the individual graph is generated. Each node in an individual graph is an arterial segment. We extract the pixel-derived features, position-based features, and topological features for each arterial segment; all features are denoted as the features for node in an individual graph, as $x_i^g$, s.t. $g \in \{1,2\}$ and $i \in \{1, \ldots, n_g\}$. For pixel-derived features, we extract the radiomics features as in our previous work [14] and [31]. For position-based features and topological features, we designed 22 hand-crafted features, as shown in Table 1.

**Table 1**. Hand-crafted features for each arterial segment

| Type | Index | Feature description |
|---|---|---|
| Pixel feature | 1 | Number of pixels in the artery segment |
| | 2 | Length of centerline |
| | 3-6 | Standard deviation, mean, the minimum and maximum radius of the artery segment |
| | 7-24 | First-order statistical radiomics features describing the distribution of pixel intensities within the arterial segments |
| | 25-48 | Gray-level co-occurrence matrix (GLCM) features describing the second-order joint probability function of an arterial segment |
| | 49-62 | Gray level dependence matrix (GLDM) features, counting the number of connected pixels within distance $\delta$ that are dependent on the center pixel |
| | 63-78 | A gray level run length matrix (GLRLM) feature quantifies gray level runs, represented by the length in the number of pixels that have the same gray level value |
| | 79-94 | Gray level size zone (GLSZM) features, counting the gray level zones in the arterial segment |
| | 95-99 | Neighboring gray-tone difference matrix (NGTDM) features, counting the difference between the gray value of a pixel and the average of its neighbors |
| Position feature | 100-103 | Weighted and absolute centers of the segment positions related to the center of the vascular tree |
| | 104-111 | Weighted and absolute positions of the two key points related to the vascular tree center |
| | 112-119 | Weighted and absolute positions of the two key points related to the artery segment center |
| Topology feature | 120-121 | Degree of the two key points |

*ii) Extracting features for vertices in the association graph.* The vertex features are generated by concatenating node features from individual graphs $G^1$ and $G^2$, as shown in Eq. 2.

$$x_{ia} = [x_i^1, x_a^2], \text{ s.t. } i \in \{1, \cdots, n_1\}, a \in \{1, \cdots, n_2\} \tag{2}$$

where [·] is the concatenation operator. For the edges in association graphs, the features are generated by concatenating features of the edges in the individual graphs, as shown in Eq. 3.

$$e_{(ia,jb)} = [e_{ij}^1, e_{ab}^2], \text{ s.t. } i, j \in \{1, \cdots, n_1\}, a, b \in \{1, \cdots, n_2\} \tag{3}$$

where $e_{ij}^g = [x_i^g, x_j^g], g \in \{1,2\}$ and $i, j \in \{1, \ldots, n_g\}$ represents features of edge in the individual graph, constituted by the concatenation of the features of two connected nodes $V_i^g$ and $V_j^g$. Then, $e_{(ia,jb)}$ indicates the features of edges in association graphs.

*iii) Feature representation learning.* We first develop a feature embedding module to embed the node and edge features in the association graph into latent representations by multi-layer perceptions (MLPs). In this study, the feature embedding is performed separately on vertices and edges, denoted as $f_{emb}^v$ and $f_{emb}^e$. Formally, the feature embedding module is defined in Eq. 4.

$$\begin{aligned} x_{ia}^A &= f_{emb}^v(x_{ia}) \\ e_{(ia,jb)}^A &= f_{emb}^e(e_{(ia,jb)}) \\ G_{emb}^A &= [f_{emb}^v(x_{ia}^A), f_{emb}^e(e_{(ia,jb)}^A)], \text{ s.t. } i,j \in \{1,\cdots,n_1\}, a,b \in \{1,\cdots,n_2\} \end{aligned} \tag{4}$$

After performing feature embedding, a GCN is employed to aggregate features from the adjacent vertices for message passing [32,33]. We adopt the message-passing neural network to aggregate features from the adjacent vertices [34], which contains a message-passing phase for feature aggregation and a readout phase for feature update. In detail, the edge convolution layer first aggregates features from the two connected vertices, and then updates its features iteratively, as shown in Eq. 5.

$$e_{(ia,jb)}^{t+1} = \phi_e([e_{(ia,jb)}^t, x_{ia}^t, x_{jb}^t]), s.t. i,j \in \{1,\cdots,n_1\}, a,b \in \{1,\cdots,n_2\}, \text{and } t \in [1,\cdots,N_{mp}] \tag{5}$$

where $\phi_e$ is the edge convolution layer implemented by MLP and $e_{(ia,jb)}^{t+1}$ becomes the updated edge features. $t$ indicates the index of the message passing, and $N_{mp}$ is the total number of message-passing steps. If $t = 1$, then $e_{(ia,jb)}^t = e_{(ia,jb)}^A$ and $x_{ia}^t = x_{ia}^A$ as defined in Eq. 4.

For each vertex, a vertex convolution layer is employed to aggregate features from the adjacence edges. The vertex convolution layer first aggregates features from the adjacent edges in the association graph and then updates its features iteratively, as shown in Eq. 6.

$$x_{ia}^{t+1} = \phi_v\left(\left[\sum_{jb \in E_{ia}} e_{(ia,jb)}^{t+1}, x_{ia}^t\right]\right), s.t. \, i,j \in \{1, \cdots, n_1\}, a, b \in \{1, \cdots, n_2\} \text{ and } t \in [1, \cdots, N_{mp}] \quad (6)$$

where $E_{ia}$ is a set containing the connected edges of vertex $x_{ia}^A$, and $\sum_{jb \in E_{ia}} \cdot$ indicates the element-wise summation of the features from the adjacent edges. $\phi_v$ is the vertex convolution layer implemented by MLPs. The per-vertex and per-edge features are computed independently, and the weights of vertex convolution layer and edge convolution layer are shared to calculate per-vertex and per-edge affinity. According to Eqs. 5 and 6, the updated vertex and edge features are denoted as $x_{ia}^{N_{mp}}$ and $e_{(ia,jb)}^{N_{mp}}$.

Unlike nature images, the visibility and anatomy between ICA images are different. Thus, the number of nodes in two individual graphs may be different. Since we assume $n_1 \leq n_2$, we manually select $G^1$ and $G^2$ so that the number of nodes in $G^1$ is smaller than that in $G^2$. The selected $G^1$ and $G^2$ are used as a pair for training the AGMN. Because the per-edge embedding, per-vertex embedding, and graph convolution layers are reused across all edges and vertices, the designed AGMN automatically supports a form of combinatorial optimization for graphs with a varying number of nodes, which is suitable and feasible for coronary artery graph matching. Thus, the proposed AGMN is independent of the input individual graphs, allowing it to perform inexact graph matching rather than the exact mapping problem for individual graphs with the same number of nodes [35].

After iteratively updating the edge features and vertex features, an MLP decoder module is employed to convert the learned feature representation to vertex classification probability, denoted as $\phi_d$. Formally, the output of the AGMN is shown in Eq. 7.

$$\hat{y}_{ia} = \phi_d\left(x_{ia}^{N_{mp}}\right), s.t. \, i \in \{1, \cdots, n_1\}, a \in \{1, \cdots, n_2\} \quad (7)$$

where $\hat{y}_{ia}$ indicates the probability of vertex $V_{ia}^A$ belonging to a positive vertex.

*iv) Vertex classification.* Graph matching is equivalent to vertex classification, so a vertex classifier is adopted to predict the matching results. According to the decoder, the matching probability between $x_i^1$ and $x_a^2$ is calculated by the decoder $\phi_d$ as $\hat{y}_{ia}$. Since each arterial branch is only matched with one arterial branch, a major probability voting strategy is employed to generate the final prediction. Formally, the vertex classification result is defined as:

$$\hat{y}_{ia} = \begin{cases} 1, if \, \text{argmax}_{k \in \{1,\ldots,n_2\}} \hat{y}_{ik} = a \\ 0, otherwise \end{cases} \quad (8)$$

In other words, the vertex $\hat{y}_{ia}$ in the vertex set $\{x_{i1}^A, x_{i2}^A, \cdots, x_{in_2}^A\}$ is selected as the positive vertex if $\hat{y}_{ia}$ has the highest probability among other vertices $\{x_{i1}^A, \cdots, x_{i,(a-1)}^A, x_{i,(a+1)}^A \cdots, x_{in_2}^A\}$.

*v) Loss function.* The node-to-node correspondence between the vertex classification results and the ground truth is used to guide the model training. The ground truth and classification results are denoted as two permutation matrix $M \in \mathbb{R}^{n_1 \times n_2}$ and $\widehat{M} \in \mathbb{R}^{n_1 \times n_2}$, where the element in $i$-th row and $a$-th column indicates the relationship between node $V_i^1 \in G^1$ and node $V_a^2 \in G^2$. The permutation loss [26] computed by the cross entropy between the predicted vertex class and the ground truth is used as an objective function, as shown in Eq. 9.

$$L_{perm} = cross\_entropy(M, \widehat{M}) = -\left(\sum_{i=1}^{n_1}\sum_{a=1}^{n_2}\left((1-\hat{y}_{ia})\log(1-y_{ia}) + \hat{y}_{ia}\log y_{ia}\right)\right) \quad (9)$$

where $y_{ia}$ is the ground truth of the vertex $x_{ia}^A$. If $y_{ia} = 1$, then the arterial segment $V_i^1$ in $G^1$ and the arterial segment $V_a^2$ in $G^2$ have the same semantic labels.

### 3.3. Training and testing

Before introducing the training and testing strategies, we first demonstrate the method to generate the labeled dataset. We manually annotated the ICA images with semantic labels for each arterial segment. Then, the semantic label was assigned to each node (i.e., each arterial segment) in the individual graph. During the creation of the database, the node correspondences between arterial segments are automatically identified based on the semantic labels, i.e. $y_{ia} = 1$ if two arterial segments have the same types. However, the main branches, such as LCX and LAD, are separated into several small branches during the individual graph generation. Then, the arterial branches with the same semantic labels may have more than one node in the individual graph. For example, in Figure 3 (d), the LAD branch has two segments due to the bifurcation points for side branch D1. These segments are matched with the LAD segments from another individual graph. If we build the ground truth of the association graph without re-naming semantic labels, a complete bipartite graph or biclique will be built; in the bipartite graph or biclique, every node of the first set is connected to every node of the second set [36]. In this example, the first LAD branch will have two matched nodes, and two vertices connecting the first LAD branch are positive vertices in the association graph. If a well-trained AGMN is obtained, the major probability voting strategy will fail to generate the final decision because these two vertices' probabilities are equal to 1.

Our designed AGMN requires the individual graph and the matching relationship to satisfy the constraint of one-to-one or one-to-zero mapping. One arterial segment or one node $V_i^1$ in the individual graph $G^1$ should only have one or zero matched node in the graph $G^2$. In this study, we annotate the arteries into several sub-classes. For example, the LAD branch in Figure 3 (d) is separated into two segments named LAD1 and LAD2. The LMA is upstream of the blood flow, and we follow this flow to assign the indices of the main and side branches sequentially. Therefore, the LCX and LAD segments are separated into several sub-segments. Each node of the first graph is only connected to one node of the second graph, and the major probability voting strategy is then usable for this task.

To train the model, at each training step, two individual graphs from the same view were randomly selected from the training set $D_{tr}$. Only left coronary arteries (LCAs) were used for model training and validation in this study. In addition, only ICAs from two regular views, left anterior oblique (LAO) and right anterior oblique (RAO), were enrolled. Because of the anatomical difference between ICAs from different views, the two selected individual graphs were from the ICAs with the same view. Since we assumed the number of nodes $n_1 \leq n_2$ in this study, we had to switch $G^1$ and $G^2$ according to the number of nodes during the training. A batch of graph pairs was randomly generated for each training iteration to accelerate the model training [37] and prevent the weights from trapping into the local minimum [38]. The association graph was built according to the two selected graphs' semantic labels. The difference between the AGMN prediction and the ground truth was used to calculate the loss and train the model, as defined in Eq. 9.

For the model testing, each individual graph in the test set $D_{te}$ is used to perform graph matching with a set of graphs. The set is denoted as the *template set, $D_{tp}$*. Cardiologists learn to read and understand the ICA in clinical practice by comparing it with the representative ICAs. When making decisions, a set of representative ICAs can be used as templates for reference. Our designed testing strategy imitates this procedure. Each individual graph from the test set is paired with the individual graph from one representative subject in the template set for graph matching. In the template set, each arterial segment is labeled for reference. Using the well-trained AGMN, the mapping relationship between unlabeled arteries in the test subject and the labeled arteries from the template subject are obtained. The vertex classification result for the test subject among the subjects in the template set is voted based on maximum voting. For example, if LMA in the test subject is matched with LMA branches from five subjects in the template set and is matched with LAD branches from two subjects in the template set, then this arterial segment in the test subject is labeled as the LMA branch.

The designed training and testing strategies of our AGMN is shown in Algorithm 2.

**Algorithm 2**. Training and testing strategies of the proposed GMN for coronary arterial semantic labeling

---

**Input**:
- $D_{tr}$: training set, contains $N_{tr}$ labeled individual graphs.
- $D_{te}$: test set, contains $N_{te}$ unlabeled individual graphs.
- $D_{tp}$: template set, contains $N_{tp}$ labeled individual graphs.
- $N$: number of training steps.
- $N_{mp}$: number of the message passing times.

**Output**: $N_{tp}$ labeled individual graphs in $D_{te}$.

**Training:**
For $iter = 1 \cdots N$ do
1. Random select two individual graphs $G^1$ and $G^2$ from $D_{tr}$ from the same view;
2. Extract features for each arterial segment in $G^1$ and $G^2$;
3. Build association graph $G^A$ and extract features using Eqs. 2 and 3;
4. Update vertex and edge features of $G^A$ using Eqs. 4 to 6 for $N_{mp}$ iterations by GCN;
5. Decode features and calculate the vertex class using Eqs. 7 and 8;
6. Calculate the loss function $L_{perm}$ defined in Eq. 9 and optimize the AGMN.

**Testing:**
For each individual graph $G_i^{te}$ ($i \in [1, \cdots, N_{te}]$) in $D_{te}$:
1. Extract features for each arterial segment in $G_i^{te}$;

For each individual graph $G_j^{tp}$ ($j \in [1, \cdots, N_{tp}]$) in $D_{tp}$:
2. Extract features for each arterial segment in $G_j^{tp}$;
3. Build the association graph $G^A$ using $G_i^{te}$ and $G_j^{tp}$;
4. Update vertex and edge features of $G^A$ using Eqs. 4 to 6 for $N_{mp}$ iterations;
5. Decode features and calculate the vertex class using Eqs. 7 and 8;
6. Assign labels for nodes in $G_i^{te}$ according to major voting among $G_j^{tp}, j \in [1, \cdots, N_{tp}]$.

---

### 3.4. Performance evaluation

The semantic labeling problem is converted into a multi-class classification problem among arterial segments. As a classification problem, the weighted accuracy (ACC), weighted precision (PRE), weighted recall (REC), and weighted F1-score (F1) are used to evaluate the model performance. We separate the LAD and LCX branches into sub-segments during the model training. However, in the evaluation process, we group the sub-segments into their original classes. The weighted ACC, SP, SN, and F1 definitions are shown in Eqs. 10 to 13.

$$ACC = \frac{1}{n}\sum_{c=1}^{C} \frac{TP_c + TN_c}{TP_c + TN_c + FN_c + FP_c} \times n_c \qquad (10)$$

$$PRE = \frac{1}{n}\sum_{c=1}^{C} \frac{TP_c}{TP_c + FP_c} \times n_c \qquad (11)$$

$$REC = \frac{1}{n}\sum_{c=1}^{C}\frac{TN_c}{TN_c + FN_c} \times n_c \qquad (12)$$

$$F1 = \frac{1}{n}\sum_{c=1}^{C}\frac{TP_c}{TP_c + \frac{1}{2}(FP_c + FN_c)} \times n_c \qquad (13)$$

where $TP_c$, $TN_c$, $FP_c$ and $FN_c$ represent the true positive arterial segment, true negative arterial segment, false positive arterial segment, and false negative arterial segment, respectively. $c$ refers to the class index of arterial segments, and $C$ is the total number of classes. $n_c$ is the number of arterial segments in class $c$ and $n$ is the total number of arterial segments.

## 4. Experiments and results

In this section, we conduct experiments to demonstrate the effectiveness of the proposed AGMN for coronary artery semantic labeling. The enrolled subjects, experimental settings, and results will be presented.

### 4.1. Dataset and enrolled subjects

In this study, we manually annotated 204 and 59 ICAs from site 1 [15] at The First Affiliated Hospital of Nanjing Medical University and site 2 at Chang Bing Show Chwan Memorial Hospital, respectively. In total, this retrospective study enrolled 263 ICA images. For site 1, subjects who received ICA from February 26, 2019, to July 18, 2019, were enrolled. The ICAs were performed using an interventional angiography system (AXIOM-Artis, Siemens, Munich) and were acquired at 15 frames/sec. The image size of ICA videos ranged from 512×512 to 864×864, and the pixel spacing ranged from 0.2 mm to 0.39 mm. For site 2, the ICAs were performed using an interventional angiography system (AlluraClarity, Philips Healthcare, Eindhoven, Netherlands) and were acquired at 3.75, 7.5, 15, and 30 frames/sec. The image size of ICA videos was 1024×1024, and the pixel spacing was 0.184 mm. Table 1 shows the number of images in each ICA view used in this study.

**Table 1.** Views and corresponding image numbers in this paper. LCA, left coronary artery. LAO, left anterior oblique; RAO, right anterior oblique.

| Site | LAO | RAO | Total |
|---|---|---|---|
| Site 1 | 55 | 149 | 204 |
| Site 2 | 23 | 36 | 59 |

For each patient, a frame that was used for anatomical structure analysis in clinical practice was selected from the view video for semantic labeling. In this study, we only focus on semantic labeling for the main branches of LMA, LAD, and LCX, and the side branches of D and OM.

### 4.2. Implementation details

We implemented our designed AGMN using TensorFlow and GraphNets [33] on an NVIDIA RTX 3090 GPU card. The thresholds used in *Algorithm 1* were set as $T_d = 1.8$ mm, $T_c = 15$ pixels, and $T_{sp} = 8$ pixels. For the 263 ICA images, we selected first $N_{tp}$ images as the labeled individual graphs for the template set, while the rest ICAs were used for a five-fold cross-validation with the stratified sampling according to the view angles of ICAs. Consequently, for each experiment, the training set contained $N_{tr} = (263 - N_{tp}) \times 0.8$ samples, and the test set contained $N_{te} = (263 - N_{tp}) \times 0.2$ samples. All images were resized to $512 \times 512$ before extracting the features. Each model was fine-tuned for 100,000 training steps using a batch size of 32. The Adam optimizer [39], with an initial learning rate of 0.0001, was employed as the optimizer. We used the exponential decay strategy in the training phase to adjust the learning rate, and we set the decay rate as 0.98 for each 2000 training steps. Each training step took 0.202 seconds, and the total training time one-fold was 5.6 hours. The hyperparameters were tuned on the test set during the cross-validation for each hyperparameter setting. The grid search settings are shown in Table 2.

**Table 2**. Hyperparameter settings in the grid search

| Hyperparameter | Search space | Description |
| --- | --- | --- |
| Number of hidden units in MLP | [16, 32, 64] | The MLP includes the feature embedding module, GCN, and feature decoder module. For each experiment, these two hyperparameters are identically set to all MLP layers. |
| Number of MLP layers | [2, 3, 4] | |
| Number of the message passing steps ($N_{mp}$) | [2, 3, 4] | The number of the message passing steps indicates the update iterations of the GCN module in GMN. |
| Number of samples in the template set ($N_{tp}$) | [27, 40, 52, 79] | 10%, 15%, 20% and 30% of the ICA images were selected using the stratified sampling according to the view angles as the template set. |

More specifically, we first fixed the hyperparameters of $N_{tp}$ as 40 (15% of the dataset) and $N_{mp}$ as 3 before we tuned the MLP layer size hyperparameters and the number of MLP layers. Then, we fixed the number of hidden units in MLP, and the number of MLP layers, then tuned the hyperparameters of $N_{tp}$ and $N_{mp}$. We selected the best parameter for each hyperparameter according to the highest average accuracy among the five-fold evaluation.

### 4.3. Experimental results of AGMN

The best performance was achieved under the settings that the number of hidden units was 64 with 4 MLP layers. The AGMN was trained using 40 samples (15%) as the template set. And the number of message-passing steps was set as 4. Under this setting, the results for each type of arterial segment are listed in Table 3.

**Table 3.** A summary of the best performance achieved by our proposed AGMN for coronary artery semantic labeling. LMA, left main artery; LAD, left descending artery; LCX, left circumflex artery; D, diagonal artery; OM obtuse margin.

| Artery type | ACC | PRE | REC | F1 |
| --- | --- | --- | --- | --- |
| LMA | 0.9956±0.0089 | 0.9911±0.0109 | 0.9956±0.0089 | 0.9933±0.0089 |
| LCX | 0.8432±0.0306 | 0.8476±0.0481 | 0.8432±0.0306 | 0.8452±0.0386 |
| LAD | 0.8046±0.0452 | 0.8256±0.0307 | 0.8046±0.0452 | 0.8143±0.0310 |
| D | 0.7956±0.0412 | 0.7536±0.0493 | 0.7956±0.0412 | 0.7736±0.0424 |
| OM | 0.7565±0.0825 | 0.7613±0.0319 | 0.7565±0.0825 | 0.7569±0.0508 |
| All | 0.8264±0.0302 | 0.8276±0.0298 | 0.8264±0.0302 | 0.8262±0.0301 |

Three examples are visualized for the graph-matching results using our proposed AGMN in Figure 5.

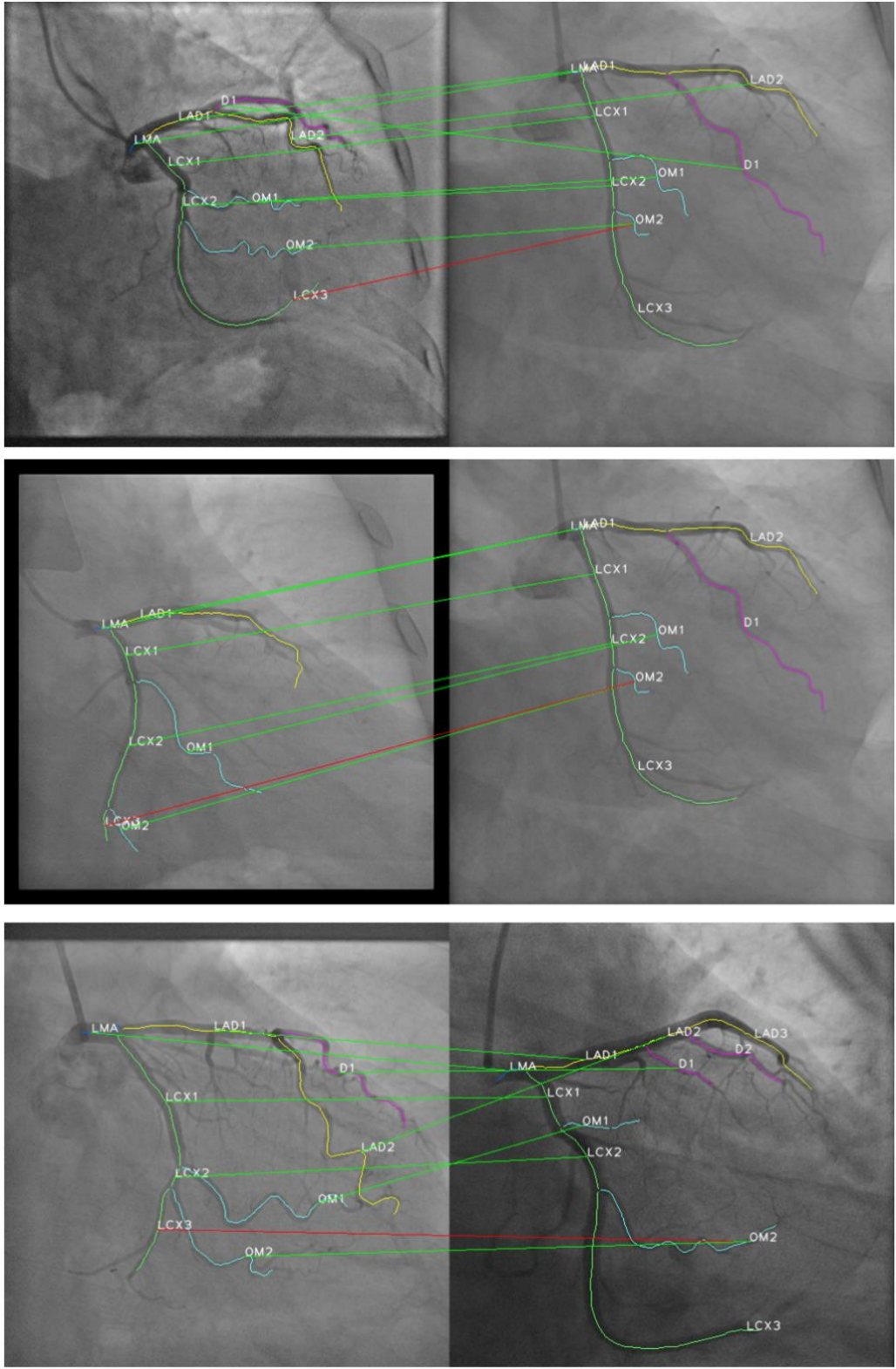

**Figure 5.** Graph matching results for three examples. The left ICAs are from the testing set, and the right ICAs are from the template set. The green line indicates a correct match, and the red line represents a wrong match. The corresponding arterial semantic labels are annotated.

## 4.4. Comparison with other coronary artery semantic labeling methods

We compared our proposed AGMN approach to four other coronary artery semantic labeling approaches. Those benchmark approaches include:

- Machine learning-based method [14]. Our previous work employed machine learning based methods to perform coronary artery semantic labeling using ICAs. Each artery is extracted using our FP-U-Net++ model and converted to vascular centerlines using step 1 to 3 in Algorithm 1. We adopted the same pipeline, extracted the same features, and employed support vector machine with radial basis function kernel as the classifier to perform arterial segment classification.
- Bi-directional tree long short-term memory (BiTreeLSTM) [6]. Wu et al. developed a tree-structured LSTM neural network for coronary artery labeling using CCTA. Arterial spatial and directions in 3D polar coordination were used as the arterial segment features. The graph-structured coronary artery was converted to a tree-structured arterial tree, and a bi-directional tree LSTM was adopted for segment classification. The bi-directional sequence input stems from LMA to side branches and from side branches to LMA. Since our ICA images are in 2D, we only extracted arterial spatial and directions in 2D polar coordination as features. We adopted the same architecture used in their paper for arterial segment classification.
- Up-to-down (UTD) and down-to-up (DTU) nets [6]. These two baseline models were implemented as the ablation study for BiTreeLSTM. Both UTD and DTU nets were implemented by a single tree LSTM. However, the UTD net used the arterial segments from the root of the coronary vascular tree, LMA, to any side branches for training, while the DTU net used the segments from the side branches to LMA for training.
- Conditional partial-residual GCN (CPR-GCN) [20]. Yang et al. developed a GCN neural network embedded with a partial residual network for coronary artery semantic labeling using CCTA. The positional features contained the arterial segment direction and position, and the imaging features were extracted by convolution neural networks (CNN) and LSTM. All features were concatenated and used for training. The output of the GCN is the segment classification result. We replaced the 3D position features with 2D features and replaced the 3D CNNs with 2D CNNs as the baseline.

The performance comparisons between the proposed AGMN and baseline models are illustrated in Table 4. We adopted five-fold cross-validation for each model and employed stratified sampling to split the samples into training and testing datasets. For baseline models, including UTD, DTU, BiTreeLSTM, and CPR-GCN, we also performed the grid search, and the models with the best performance were used for comparison.

**Table 4.** Comparisons between baseline methods and our proposed AGMN for coronary artery semantic labeling using our ICA dataset. The means and standard deviations of the accuracy, precision, recall, and F1-scores among the five folds are presented. The bold texts indicate they achieved the best performance in their corresponding evaluation metrics.

| Method | Metric | LMA | LAD | LCX | D | OM | All |
|---|---|---|---|---|---|---|---|
| Machine learning [14] | ACC | 0.9925±0.0151 | 0.6331±0.0540 | 0.6388±0.0390 | 0.6147±0.0410 | 0.5907±0.0419 | 0.6651±0.0080 |
|  | PRE | 0.9778±0.0071 | 0.6586±0.0174 | 0.6375±0.0378 | 0.5554±0.0101 | 0.6278±0.0261 | 0.6679±0.0081 |
|  | REC | 0.9925±0.0151 | 0.6331±0.0540 | 0.6388±0.0390 | 0.6147±0.0410 | 0.5907±0.0419 | 0.6651±0.0080 |
|  | F1 | 0.9850±0.0076 | 0.6437±0.0213 | 0.6360±0.0087 | 0.5832±0.0234 | 0.6071±0.0183 | 0.6646±0.0077 |
| UTD net [6] | ACC | 1.0000±0.0000 | 0.8828±0.0136 | 0.9245±0.0235 | 0.0000±0.0000 | 0.0000±0.0000 | 0.6182±0.0094 |
|  | PRE | 1.0000±0.0000 | 0.7927±0.0167 | 0.4629±0.0106 | 0.0000±0.0000 | 0.0000±0.0000 | 0.4562±0.0069 |
|  | REC | 1.0000±0.0000 | 0.8828±0.0136 | 0.9245±0.0235 | 0.0000±0.0000 | 0.0000±0.0000 | 0.6182±0.0094 |
|  | F1 | 1.0000±0.0000 | 0.8353±0.0142 | 0.6169±0.0143 | 0.0000±0.0000 | 0.0000±0.0000 | 0.5135±0.0075 |
| DTU net [6] | ACC | 0.0000±0.0000 | **1.0000±0.0000** | 0.7359±0.0194 | 0.0000±0.0000 | 0.3575±0.2923 | 0.5450±0.0516 |
|  | PRE | 0.0000±0.0000 | 0.4284±0.0044 | 0.6795±0.1010 | 0.0000±0.0000 | 0.5924±0.4837 | 0.4264±0.1212 |
|  | REC | 0.0000±0.0000 | **1.0000±0.0000** | 0.7359±0.0194 | 0.0000±0.0000 | 0.3575±0.2923 | 0.5450±0.0516 |
|  | F1 | 0.0000±0.0000 | 0.5998±0.0043 | 0.7018±0.0532 | 0.0000±0.0000 | 0.4458±0.3643 | 0.4491±0.0838 |
| BiTreeLSTM [6] | ACC | 1.0000±0.0000 | 0.8845±0.0150 | 0.9871±0.0120 | 0.0000±0.0000 | 0.5981±0.0165 | 0.7492±0.0085 |
|  | PRE | 1.0000±0.0000 | **0.8562±0.0190** | 0.5853±0.0099 | 0.0000±0.0000 | 0.9808±0.0122 | 0.6927±0.0074 |
|  | REC | 1.0000±0.0000 | 0.8845±0.0150 | **0.9871±0.0120** | 0.0000±0.0000 | 0.5981±0.0165 | 0.7492±0.0085 |
|  | F1 | 1.0000±0.0000 | **0.8699±0.0101** | 0.7348±0.0093 | 0.0000±0.0000 | 0.7429±0.0141 | 0.6967±0.0085 |
| CPR-GCN [20] | ACC | 0.5361±0.2996 | 0.5319±0.1239 | 0.5072±0.1447 | 0.0624±0.0953 | 0.5341±0.3045 | 0.4581±0.0536 |
|  | PRE | 0.6208±0.3240 | 0.5675±0.0540 | 0.3964±0.0570 | 0.2802±0.3727 | 0.3821±0.0139 | 0.4463±0.1075 |
|  | REC | 0.5361±0.2996 | 0.5319±0.1239 | 0.5072±0.1447 | 0.0624±0.0953 | 0.5341±0.3045 | 0.4581±0.0536 |
|  | F1 | 0.5698±0.3026 | 0.5455±0.0899 | 0.4353±0.0632 | 0.0742±0.0957 | 0.3924±0.1660 | 0.4192±0.0661 |
| Our AGMN | ACC | **0.9956±0.0089** | 0.8432±0.0306 | **0.8046±0.0452** | **0.7956±0.0412** | **0.7565±0.0825** | **0.8264±0.0302** |
|  | PRE | **0.9911±0.0109** | 0.8476±0.0481 | **0.8256±0.0307** | **0.7536±0.0493** | **0.7613±0.0319** | **0.8276±0.0298** |
|  | REC | **0.9956±0.0089** | 0.8432±0.0306 | 0.8046±0.0452 | **0.7956±0.0412** | **0.7565±0.0825** | **0.8264±0.0302** |
|  | F1 | **0.9933±0.0089** | 0.8452±0.0386 | **0.8143±0.0310** | **0.7736±0.0424** | **0.7569±0.0508** | **0.8262±0.0301** |

According to Table 4, our AGMN achieved the highest average accuracy of 0.8264, average precision of 0.8276, average recall of 0.8264, and average F1-score of 0.8262 among all types of coronary arteries, which outperformed other baseline models significantly.

All models achieved high performance on main branches prediction, except for DTU net. The UTD and BiTreeLSTM performed the perfect classification on LMA; the accuracy was 100%. Because LMA was required to be manually assigned to build the tree-structured arteries, it did not reflect its actual performance. The input sequence of UTD and BiTreeLSTM stems from LMA to the side branches. For the LAD branch, the BiTreeLSTM achieved a higher precision and F1-score than the proposed AGMN due to the correct classification of LMA. Since LAD is connected to LMA directly, so the classification task is relaxed. For the LCX branch, the BiTreeLSTM achieved a higher recall than the proposed AGMN because of the prior knowledge of the LMA branch.

Our AGMN outperformed other baselines with a large margin for the side branches. By comparing the similarities of side branches between different individual graphs, the AGMN can differentiate D branches and OM branches. In addition, the D branches are connected with LAD, and OM branches are connected with LCX; the performance of side branch classification is guaranteed since the AGMN has achieved a high classification performance of LAD and LCX branches.

### 4.5. Feature importance

In this study, we designed 121 hand-craft features, including pixel features, positional features, and topological features. A leave-one-out technique [40] is adopted to identify the feature significance. A feature is significant if the performance of semantic labeling decreases significantly when this feature is replaced by zero. By ranking the accuracy drops, the importance of the feature is obtained. We set the same hyperparameters demonstrated in section 4.3. We compared the averaged the accuracy changes among five-folds between using the original dataset and the corrupted datasets, as shown in Figure 6.

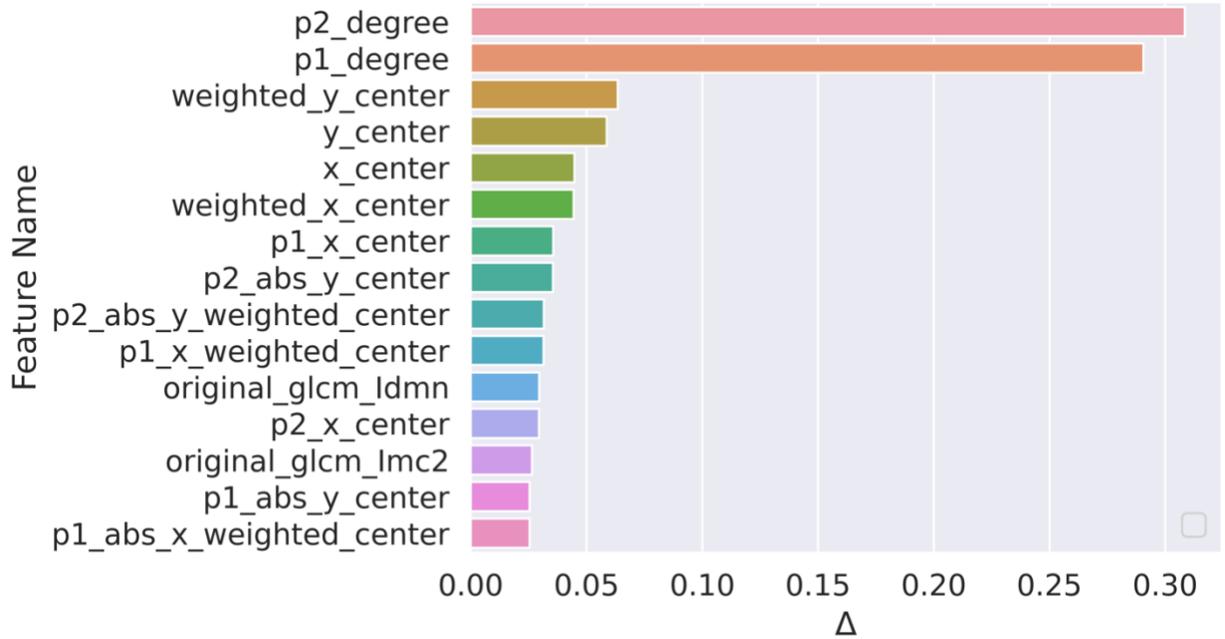

**Figure 6**. Feature importance ranking for classifying coronary arterial segments. Feature significance was determined by the accuracy drops between using raw features and zero-filled features. The vertical axis indicates the feature names, while the horizontal axis indicates the drops in accuracy.

The top 15 features with the highest accuracy changes are shown in Figure 6. Among these 15 features, 2 are topological features, 2 are pixel-wise features and the other 11 are positional features. For the topological features, *p1 degree* and *p2 degree* indicate the degrees of the two endpoints of an artery segment. If we set these two features as zeros, then the accuracy of coronary artery semantic labeling dropped about 30%, indicating the convincible importance of these two features. For the pixel features, *original_glcm_ldmn* and *original_glcm_lmc2* are the features calculated by the Gray Level Co-occurrence Matrix, which describes the second-order joint probability function of the region of the arterial segments. The remaining 11 features belong to hand-craft positional features, representing the weighted or absolute coordinates of the center pixels within the arterial segments. Due to the anatomical structure of the coronary artery, the position of the artery is important for classification.

### 4.6. Data attack

The proposed AGMN was trained and evaluated based only on the 'ideal' individual graphs. However, even though our previous coronary artery binary segmentation model [12] has achieved the Dice similarity coefficient of 0.8899, we cannot guarantee that it would generate satisfactory arterial contours for all ICAs. To test the robustness of the designed model, we created the corrupted datasets by randomly removing parts of arterial segments from the ICAs in the test set. The removed arterial segment must contain one endpoint to generate a connected graph. Otherwise, if the arterial segment contains 2 bifurcation points and is removed, the individual arterial graph would be split into two individual graphs as well as the vascular tree would be separated. In this situation, human intervention is required.

Using the corrupted dataset, we compared the AGMN with the graph- or tree-based baseline models, including BiTreeLSTM and CPR-GCN. We compared the performance drops using the corrupted dataset by randomly removing 5%, 7.5%, 10%, 12.5%, 15%, 17.5%, and 20% arterial segments. The ACC, PREC, REC, and F1 and their changes using different corrupted datasets are shown in Figure 7.

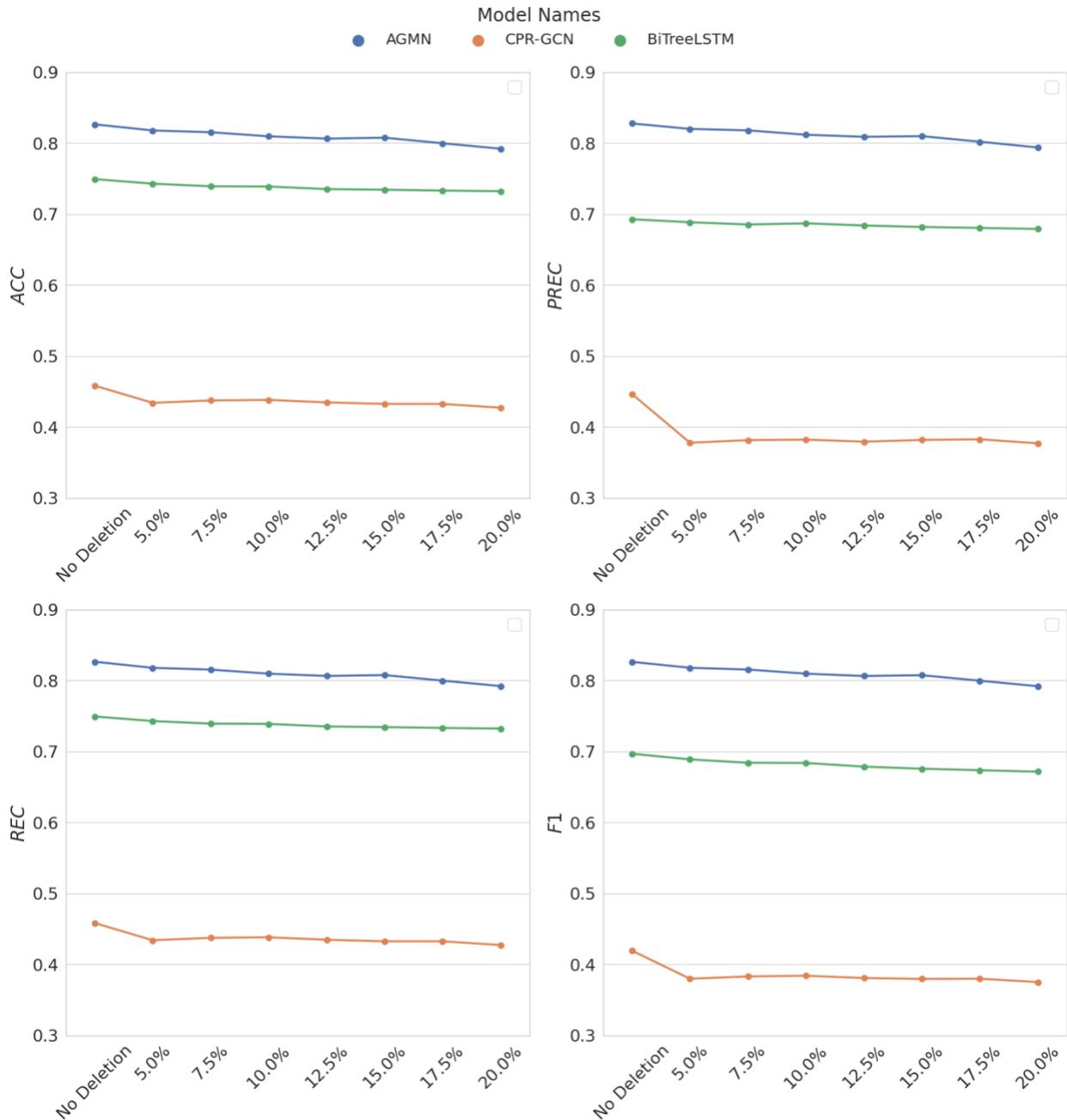

**Figure 7.** The achieved ACC, PREC, REC, and F1 of the proposed AGMN, CPR-GCN, and BiTreeLSTM using different corrupted datasets. The horizontal axis indicates the probability of deleting an artery segment randomly.

The results demonstrated that our AGMN was robust since the accuracy was above 0.79, even using the corrupted datasets with 20% of the arterial segments removed. However, the accuracy changes of AGMN were more significant than those of BiTreeLSTM. According to Table 4, we observed that the BiTreeLSTM achieved a low performance on side branch classification. When generating the corrupted datasets, parts of the side branches were removed, which didn't affect the overall performance significantly. We didn't remove the LMA branches when testing the BiTreeLSTM using the corrupted datasets. The LMA is the root of the arterial tree. If we removed the LMA branch, the arterial tree would be separated into two sub-trees. The tree-based model, BiTreeLSTM, failed to perform artery semantic labeling when using the data without LMA, which is the root of the vascular tree. In addition, our proposed AGMN achieved the highest ACCs, PRECs, RECs, and F1s using different corrupted datasets than other baseline models, indicating its robustness and powerful performance.

## 5. Limitations and future work

The limitation of the proposed AGMN is that during the prediction, the graph matching procedures are required to be performed between the test subject and every subject in the template set. If we employ fewer subjects in the template set, the prediction time would be reduced. In the future, graph clustering will be used to select the most representative subjects in each cluster and then construct the template set to accelerate the prediction.

We use hand-crafted features as the pixel-level features to reduce the training time and model complexity. However, the feature representation capability is limited compared to CNN-extracted features. In the future, a light-weight deep learning-based method is recommended to automatically extract the pixel features for each segment rather than the hand-crafted radiomics features.

## 6. Conclusion

In this paper, we developed and validated a new algorithm for coronary artery semantic labeling on ICAs with high accuracy, interpretability, and robustness. A new workflow for the individual graph generation and the association graph-based graph matching network was proposed. The association graph-based approach, per-node, and per-edge feature representation learning network performed inexact graph matching. The experimental results showed that our AGMN achieved the best performance and significantly outperformed existing approaches. By analyzing the feature importance, the interpretability of AGMN is guaranteed. Our AGMN still performed highly in the data attack experiments, even using corrupted datasets.

**Credit authorship contribution statement**

Chen Zhao: Conceptualization, methodology, coding, manuscript writing.

Zhihui Xu: Data management and clinical validation.

Jingfeng Jiang: Methodology and manuscript writing.

Michele Esposito: Clinical validation and manuscript writing.

Drew Pienta: Methodology and manuscript writing.

Guang-Uei Hung: Data management and manuscript writing.

Weihua Zhou: Supervision, project administration, funding acquisition, manuscript writing, review.

**Declaration of Competing Interest**

The authors declare no conflicts of interest.

**Acknowledgment**

This research was supported by a new faculty startup grant from Michigan Technological University Institute of Computing and Cybersystems (PI: Weihua Zhou). It was also supported in part by a research seed fund from Michigan Technological University Health Research Institute and an NIH grant (U19AG055373).

# Reference

[1] Okrainec K, Banerjee DK, Eisenberg MJ. Coronary artery disease in the developing world. American heart journal. Elsevier; 2004;148(1):7–15.

[2] Benjamin EJ, Muntner P, Alonso A, Bittencourt MS, Callaway CW, Carson AP, Chamberlain AM, Chang AR, Cheng S, Das SR. Heart disease and stroke statistics—2019 update: a report from the American Heart Association. Circulation. Am Heart Assoc; 2019;139(10):e56–e528.

[3] Boden WE, O'Rourke RA, Teo KK, Hartigan PM, Maron DJ, Kostuk WJ, Knudtson M, Dada M, Casperson P, Harris CL, Chaitman BR, Shaw L, Gosselin G, Nawaz S, Title LM, Gau G, Blaustein AS, Booth DC, Bates ER, Spertus JA, Berman DS, Mancini GBJ, Weintraub WS. Optimal Medical Therapy with or without PCI for Stable Coronary Disease. N Engl J Med. 2007 Apr 12;356(15):1503–1516.

[4] Miller DC, Stinson EB, Rossiter SJ, Oyer PE, Reitz BA, Shumway NE. Impact of simultaneous myocardial revascularization on operative risk, functional result, and survival following mitral valve replacement. Surgery. 1978;84(6):848–857.

[5] Li Z, Zhang Y, Liu G, Shao H, Li W, Tang X. A robust coronary artery identification and centerline extraction method in angiographies. Biomedical Signal Processing and Control. Elsevier; 2015;16:1–8.

[6] Wu D, Wang X, Bai J, Xu X, Ouyang B, Li Y, Zhang H, Song Q, Cao K, Yin Y. Automated anatomical labeling of coronary arteries via bidirectional tree LSTMs. Int J CARS. 2019 Feb;14(2):271–280.

[7] Gifani P, Behnam H, Shalbaf A, Sani ZA. Automatic detection of end-diastole and end-systole from echocardiography images using manifold learning. Physiological measurement. IOP Publishing; 2010;31(9):1091.

[8] Esposito A, Gallone G, Palmisano A, Marchitelli L, Catapano F, Francone M. The current landscape of imaging recommendations in cardiovascular clinical guidelines: toward an imaging-guided precision medicine. Radiol Med. 2020 Nov;125(11):1013–1023. PMCID: PMC7593299

[9] Parikh NI, Honeycutt EF, Roe MT, Neely M, Rosenthal EJ, Mittleman MA, Carrozza JP, Ho KKL. Left and Codominant Coronary Artery Circulations Are Associated With Higher In-Hospital Mortality Among Patients Undergoing Percutaneous Coronary Intervention for Acute Coronary Syndromes: Report From the National Cardiovascular Database Cath Percutaneous Coronary Intervention (CathPCI) Registry. Circ: Cardiovascular Quality and Outcomes. 2012 Nov;5(6):775–782.

[10] Yang G, Broersen A, Petr R, Kitslaar P, de Graaf MA, Bax JJ, Reiber JHC, Dijkstra J. Automatic coronary artery tree labeling in coronary computed tomographic angiography datasets. 2011 Computing in Cardiology. 2011. p. 109–112.

[11] Gu S, Wang Z, Siegfried JM, Wilson D, Bigbee WL, Pu J. Automated lobe-based airway labeling. International journal of biomedical imaging. Hindawi; 2012;2012.

[12] Zhang H, Gao Z, Zhang D, Hau WK, Zhang H. Progressive Perception Learning for Main Coronary Segmentation in X-ray Angiography. IEEE Transactions on Medical Imaging. 2022;1–1.

[13] Zhai M, Du T, Yang R, Zhang H. Coronary Artery Vascular Segmentation on Limited Data via Pseudo-Precise Label. 2019 41st Annual International Conference of the IEEE Engineering in Medicine and Biology Society (EMBC). 2019. p. 816–819.

[14] Zhao C, Bober R, Tang H, Tang J, Dong M, Zhang C, He Z, Esposito M, Xu Z, Zhou W. Semantic Segmentation to Extract Coronary Arteries in Invasive Coronary Angiograms. Journal of Advances in Applied & Computational Mathematics. 2022 May 24;9:76–85.

[15] Zhao C, Vij A, Malhotra S, Tang J, Tang H, Pienta D, Xu Z, Zhou W. Automatic extraction and stenosis evaluation of coronary arteries in invasive coronary angiograms. Computers in Biology and Medicine. 2021;136:104667.

[16] Yuan X, Shi J, Gu L. A review of deep learning methods for semantic segmentation of remote sensing imagery. Expert Systems with Applications. 2021 May;169:114417.

[17] Xian Z, Wang X, Yan S, Yang D, Chen J, Peng C. Main Coronary Vessel Segmentation Using Deep Learning in Smart Medical. Huang C, editor. Mathematical Problems in Engineering. 2020 Oct 21;2020:1–9.

[18] Silva JL, Menezes MN, Rodrigues T, Silva B, Pinto FJ, Oliveira AL. Encoder-Decoder Architectures for Clinically Relevant Coronary Artery Segmentation [Internet]. arXiv; 2021. Available from: http://arxiv.org/abs/2106.11447

[19] Cao Q, Broersen A, de Graaf MA, Kitslaar PH, Yang G, Scholte AJ, Lelieveldt BPF, Reiber JHC, Dijkstra J. Automatic identification of coronary tree anatomy in coronary computed tomography angiography. Int J Cardiovasc Imaging. 2017 Nov;33(11):1809–1819.

[20] Yang H, Zhen X, Chi Y, Zhang L, Hua X-S. CPR-GCN: Conditional Partial-Residual Graph Convolutional Network in Automated Anatomical Labeling of Coronary Arteries. 2020 IEEE/CVF Conference on Computer Vision and Pattern Recognition (CVPR) [Internet]. Seattle, WA, USA: IEEE; 2020. p. 3802–3810. Available from: https://ieeexplore.ieee.org/document/9156834/

[21] Dijkman R, Dumas M, García-Bañuelos L. Graph matching algorithms for business process model similarity search. International conference on business process management. Springer; 2009. p. 48–63.

[22] Bunke H. On a relation between graph edit distance and maximum common subgraph. Pattern recognition letters. Elsevier; 1997;18(8):689–694.

[23] Lawler EL. The quadratic assignment problem. Management science. INFORMS; 1963;9(4):586–599.

[24] Scarselli F, Gori M, Ah Chung Tsoi, Hagenbuchner M, Monfardini G. The Graph Neural Network Model. IEEE Trans Neural Netw. 2009 Jan;20(1):61–80.

[25] Nowak A, Villar S, Bandeira AS, Bruna J. Revised Note on Learning Quadratic Assignment with Graph Neural Networks. 2018 IEEE Data Science Workshop (DSW) [Internet]. Lausanne, Switzerland: IEEE; 2018. p. 1–5. Available from: https://ieeexplore.ieee.org/document/8439919/

[26] Wang R, Yan J, Yang X. Learning Combinatorial Embedding Networks for Deep Graph Matching. 2019 IEEE/CVF International Conference on Computer Vision (ICCV) [Internet]. Seoul, Korea (South): IEEE; 2019. p. 3056–3065. Available from: https://ieeexplore.ieee.org/document/9010940/

[27] Riesen K, Jiang X, Bunke H. Exact and Inexact Graph Matching: Methodology and Applications. In: Aggarwal CC, Wang H, editors. Managing and Mining Graph Data [Internet]. Boston, MA: Springer US; 2010. p. 217–247. Available from: http://link.springer.com/10.1007/978-1-4419-6045-0_7

[28] Dashtbozorg B, Mendonça AM, Campilho A. An automatic graph-based approach for artery/vein classification in retinal images. IEEE Transactions on Image Processing. IEEE; 2013;23(3):1073–1083.

[29] Zhou F, De la Torre F. Factorized Graph Matching. IEEE Trans Pattern Anal Mach Intell. 2016 Sep 1;38(9):1774–1789.

[30] Loiola EM, de Abreu NMM, Boaventura-Netto PO, Hahn P, Querido T. A survey for the quadratic assignment problem. European journal of operational research. Elsevier; 2007;176(2):657–690.

[31] Zhao C, Xu Y, He Z, Tang J, Zhang Y, Han J, Shi Y, Zhou W. Lung segmentation and automatic detection of COVID-19 using radiomic features from chest CT images. Pattern Recognition. Elsevier; 2021;119:108071.

[32] Bianchi FM, Grattarola D, Alippi C. Spectral Clustering with Graph Neural Networks for Graph Pooling. arXiv:190700481 [cs, stat] [Internet]. 2020 Dec 29; Available from: http://arxiv.org/abs/1907.00481

[33] Battaglia PW, Hamrick JB, Bapst V, Sanchez-Gonzalez A, Zambaldi V, Malinowski M, Tacchetti A, Raposo D, Santoro A, Faulkner R, Gulcehre C, Song F, Ballard A, Gilmer J, Dahl G, Vaswani A, Allen K, Nash C, Langston V, Dyer C, Heess N, Wierstra D, Kohli P, Botvinick M, Vinyals O, Li Y, Pascanu R. Relational inductive biases, deep learning, and graph networks [Internet]. arXiv; 2018. Available from: http://arxiv.org/abs/1806.01261

[34] Gilmer J, Schoenholz SS, Riley PF, Vinyals O, Dahl GE. Neural Message Passing for Quantum Chemistry [Internet]. arXiv; 2017. Available from: http://arxiv.org/abs/1704.01212

[35] Wang T, Liu H, Li Y, Jin Y, Hou X, Ling H. Learning Combinatorial Solver for Graph Matching. 2020 IEEE/CVF Conference on Computer Vision and Pattern Recognition (CVPR) [Internet]. Seattle, WA, USA: IEEE; 2020. p. 7565–7574. Available from: https://ieeexplore.ieee.org/document/9156795/

[36] Bondy JA, Murty USR. Graph theory with applications. Macmillan London; 1976.

[37] Bottou L. Large-scale machine learning with stochastic gradient descent. Proceedings of COMPSTAT'2010. Springer; 2010. p. 177–186.

[38] Ge R, Huang F, Jin C, Yuan Y. Escaping from saddle points—online stochastic gradient for tensor decomposition. Conference on learning theory. PMLR; 2015. p. 797–842.

[39] Kingma DP, Ba J. Adam: A method for stochastic optimization. arXiv preprint arXiv:14126980. 2014;

[40] Wang T, Shao W, Huang Z, Tang H, Zhang J, Ding Z, Huang K. MOGONET integrates multi-omics data using graph convolutional networks allowing patient classification and biomarker identification. Nat Commun. 2021 December;12(1):3445.